\documentclass[11pt,a4paper]{article}
\usepackage{mtsummit2019}
\usepackage{times}
\usepackage{url}
\usepackage{latexsym}
\usepackage{adjustbox}

\usepackage{multirow}

\usepackage{arabtex}
\usepackage[T1]{fontenc}
\usepackage{epstopdf}
\usepackage[utf8]{inputenc}

\newcommand{\hide}[1]{}

\newcommand{\ARJAcorpus}{\sc TUFSMedia.Arabic}
\newcommand{\MaiSMT}{\sc JaPostEdit}
\newcommand{\GoSMT}{\sc TufsSMT}
\newcommand{\GTNMT}{\sc GoogNMT}
\newcommand{\train}{\sc Train}
\newcommand{\tune}{\sc Tune}
\newcommand{\devtest}{\sc Dev}
\newcommand{\blind}{\sc Test}

\usepackage{CJKutf8}  
\newcommand{\jap}[1]{\begin{CJK}{UTF8}{min}#1\end{CJK}}
\usepackage[overlap, CJK]{ruby}  
\usepackage{CJKulem}  
\title{Simple Automatic Post-editing for Arabic-Japanese Machine Translation}

\author{Ella Noll, Mai Oudah and Nizar Habash\\
Computational Approaches to Modeling Language Lab\\
New York University Abu Dhabi\\
United Arab Emirates\\
  {\tt \{ella.noll,mai.oudah,nizar.habash\}@nyu.edu}}

\date{}

\begin{document}
\maketitle
\begin{abstract}  
A common  bottleneck for developing machine translation (MT) systems for some language pairs is the lack of direct parallel translation data sets, in general and in certain domains. 
Alternative solutions such as zero-shot models or pivoting techniques are successful in getting a strong baseline, but are often below the more supported language-pair systems. 
In this paper, we focus on Arabic-Japanese machine translation, a less studied language pair; and  we work with a unique parallel corpus of Arabic news articles  that were manually translated  to Japanese. 
We use this parallel corpus to adapt a state-of-the-art  domain/genre agnostic neural MT system via a simple automatic post-editing technique.  Our results and detailed  analysis suggest that this approach is quite viable for less supported language pairs in specific domains.
\end{list}
\end{abstract}
\setarab
\novocalize


\section{Introduction} 
Machine Translation (MT) research has made impressive strides in the last two decades.
However this success has not spread equally to all language pairs, with some language pairs receiving a lot more attention in terms of research, and resource and system development.  A bottleneck for some language pairs is the lack of direct parallel translation data sets. This issue has been addressed through alternative solutions such as zero-shot models  \cite{Johnson:2016:googles} or pivoting techniques \cite{Utiyama:2007:comparison,Habash:2009:improving,Liu:2019:pivot}. Although such methods can be successful in getting a strong baseline, they are often below the more supported language-pair systems. 
A related challenge is that available data for pivot/zero-shot techniques may be different from the specific genres/domain a user may be interested in. Furthermore, the MT output  often carries features of the source language \cite{Volansky:2015:features}  that may be hard to model in a pivot or zero shot system because of the interaction effect of other languages.
These are problems in general for MT, but they are exacerbated in the context of already limited resources.

In this paper, we present results from a simple automatic post-editing system for Arabic-Japanese MT that exploits a corpus of Arabic-Japanese news articles. Our main contribution is in working with a less studied  pair of languages that are far more  different from each other than the typically studied language pairs in automatic post editing.  
Our results improve over a very strong (but domain/genre agnostic) state-of-the-art system.
Our linguistic analysis provides some insights in the type of changes made by the system.

Next, we present some related work (Section~\ref{related}), followed by relevant linguistic facts about Arabic and Japanese (Section~\ref{facts}), and an analysis of Arabic-Japanese MT errors (Section~\ref{errors}).  We then present our approach ( Section~\ref{postedit}), discuss experimental settings and results (Section~\ref{experiments}), and present an error analysis with examples (Section~\ref{analysis}).


\section{Related Work}
\label{related}
The systems we evaluate in this paper for Arabic-to-Japanese translation combine elements of three meta approaches to address limited resources: automatic post-editing, pivoting and domain adaptation.  The best result we have is for a system that is built through pivoting Arabic-to-Japanese through English, and then automatically post-edited using a corpus from a specific domain.

\subsection{Automatic Post-editing} 
Research on automatic post-editing (APE) aims to develop automatic methods for correcting errors produced by an unknown MT system \cite{chatterjee-EtAl:2018:WMT}.  
Though Japanese-specific work in the area of automatic post-editing (APE) is minimal, there have been a number of studies conducted on other languages. 
Most of this work has used English as one of the languages in its initial MT language-pair which is unexpected given the abundance of parallel-data available.

One of the earliest  reported results on APE of MT outputs are from \newcite{simard2007statistical}  who successfully performed statistical post-editing on rule-based MT outputs for English-French and French-English MT. 
 %
 \newcite{bechara2011statistical}  
 used the output of a first-stage English-French Statistical MT (SMT) system to train a monolingual second stage system, with the objective of discovering whether, and to what extent, SMT technology can be used to post-edit itself. Their results showed an improvement of around two BLEU points for all thresholds for the French-English translations, but no improvement for translations of the other direction (English-French). 
%
\newcite{rosa2013deepfix} 
attempted to correct errors in a verb-noun valency (the way in which verbs and their arguments are used together) using deep syntactic analysis and a simple probabilistic model of valency for the English-Czech translation pair. 

 \newcite{chatterjee2015acl} conducted a systematic comparison between the APE methods by \newcite{simard2007statistical} and \newcite{bechara2011statistical}. The comparison is done under the same conditions with respect to data and evaluation settings, examining six language pairs having English as the source language and Czech, German, Spanish, French, Italian and Polish as the target languages. The results suggest that considering the source words in the process of APE training as done in \newcite{bechara2011statistical} can help recovering some adequacy translation errors.

In the last four years and this year (2019), the conference on MT (WMT) has run yearly shared tasks on APE \cite{chatterjee-EtAl:2018:WMT}.  The language pairs the shared task worked on are   English-Spanish \cite{pal2015usaar} English-German \cite{chatterjee2016fbk}, and German-English \cite{chatterjee2017multi}. In 2019 English-Russian was added. The progress on APE research has been dominated by neural approaches.  In this paper we opt to use Statistical MT (SMT) post-editing applied to SMT and NMT systems. We plan to explore neural models for post-editing in the future.


\subsection{MT Pivoting} 

A commonly used solution for addressing the lack of parallel data for a certain language pair is to pivot through a third (pivot or bridge) language  that has enough shared parallel data with the two languages of interest \cite{Hajic:2000:machine,Utiyama:2007:comparison,Wu:2007:pivot,Bertoldi:2008:Phrase,Habash:2009:improving,Koehn:2009:462,Liu:2019:pivot}.
Pivoting has been shown useful for closely related languages
\cite{Hajic:2000:machine} as well as unrelated languages
\cite{Habash:2009:improving,Liu:2019:pivot}.  
English is the most commonly used pivot language simply because systems/corpora to and from English are typically more available than other possible pivots, although this is not exclusive --- \cite{Liu:2019:pivot} translated Japanese patent text to English through Chinese as a pivot.
The simplest pivoting strategy is sentence pivoting, where we build two independent systems and pipeline them.  This technique is still used by GoogleTranslate for some language pairs.
%
%
%
Phrase pivoting joins the phrase tables from two phrase-based MT systems to create a single table \cite{Utiyama:2007:comparison,Wu:2007:pivot,ElKholy:2013:language}.  The shift in the field towards neural MT and the introduction of zero-shot methods capture the pivoting/bridging intuition but in a different approach  \cite{Johnson:2016:googles}.

\subsection{MT Domain Adaptation} 

In the more specific context where parallel data exists for a language pair in general, but less so for the same language pair in a particular domain, {\it domain adaptation} techniques are used to extend existing systems or data sets 
\cite{koehn2007experiments,isabelle2007domain,Bertoldi:2009:DAS:1626431.1626468,luong2015stanford,farajian2017neural,etchegoyhen-etal-2018-evaluating}.  
The approaches vary in the degree of intrusion within an existing system: from focusing on data selection and retrain from scratch \cite{koehn2007experiments}; to a post-editing like approach that starts with a general-domain MT system and pipelines it with another MT system that post-edits the output of the first system \cite{isabelle2007domain}.


 \begin{table*}[t]

 \begin{center}
 \scalebox{0.73}{
\begin{tabular}{l|c|c|c}
& \bf Arabic & \bf English  & \bf Japanese  \\\hline\hline
\bf Orthography & Abjad & Alphabet & Logographs + Syllabary \\\hline\hline
\bf Morphology & Rich & Poor & Rich formality, Poor inflection \\\hline\hline

\bf Verbal      & Verb Subject Object PP& Subject Verb Object PP& Subject  Object PP V\\\cline{2-4}

\bf Sentence & <A^strY `ly bytA fy .twkyw>  &  & \jap{アリが東京で家を買った}   \\
\bf&   (A\$trY Ely bytA fy Twkyw)&  &   (Ari ga Tōkyō de ie o katta)\\
 & {\it bought Ali a-house in Tokyo} & Ali bought  a house in Tokyo& {\it Ali [subj] Tokyo [loc] house [obj] bought }\\\hline\hline

\bf Copular  & Subject Predicate & Subject {\it be} Predicate & Subject Predicate {\it be}\\\cline{2-4}

\bf Sentence  & <Alklb 'aswad> (Alklb Aswad) &  & \jap{犬 は 黒 です}(inu wa kuro desu) \\
& {\it the-dog black} & The dog is black & {\it dog [topic] black is (formal)} \\\hline \hline

\bf Prep Noun & Prep Noun & Prep Noun & Noun Prep \\\cline{2-4}
    & <fy AlyAbAn> (fy AlyAbAn) &  & \jap{日本に}(nihon ni)\\
    & {\it in the-Japan} & In Japan & {\it Japan in}\\    \hline\hline

\bf Adjectival  & Noun Adjective & Adjective Noun & Adjective Noun \\\cline{2-4}
 \bf Modifier  & <klb 'aswad> (kalb Aswad) &  & \jap{黒い 犬} (kuroi inu) \\
 & {\it dog black} & A black dog & {\it black dog} \\ \hline\hline

\bf Possessive  & Noun Possessor & Noun of Possessor | Possessor's Noun  & Possessor \jap{の} ({\it no}) Noun \\\cline{2-4}
\bf Modifier & { <byt `ly>} (byt Ely) && \jap{アリ の 家} (Ari no ie) \\
 & {\it house Ali} &  The house of Ali  | Ali's house  & {\it Ali [poss] house} \\\hline\hline

\bf Relative    & Noun RelClause & Noun RelClause  &  RelClause Noun \\\cline{2-4}
 \bf Modifier & <Albyt Al_dy A^strAh `ly Alywm>  &   & \jap{アリ が 今日 買った 家}  \\ 
 &  (Albyt All*y A\$trAh Ely Alywm) &    &  (Ari ga kyō katta ie)\\
 & {\it the-house that bought-it Ali today} &The house that Ali bought today & {\it Ali [subj] today bought house}\\ \hline

\end{tabular}
}
\end{center}
\caption{\label{tab:ar-en-ja-table} Dimensions of variation with examples across Arabic, English and Japanese.  All examples are parallel translations.}

\end{table*}

\subsection{Arabic-Japanese MT Resources}
\newcite{Inoue:2018:parallel} presented a parallel corpus of Arabic--Japanese news articles aligned for use in MT.
The corpus text base came from an ongoing project at Tokyo University of Foreign Studies (TUFS) entitled TUFS Media Project,\footnote{\url{http://www.el.tufs.ac.jp/tufsmedia/}} which produces translated news articles in 
eight languages (Arabic, Bengali, Burmese, Indonesian, Persian, Turkish, Urdu, and Vietnamese).
The Arabic-Japanese corpus,\footnote{\url{http://el.tufs.ac.jp/tufsmedia-corpus/}} (henceforth, {\ARJAcorpus})
consists of 64,488 sentence pairs (2.4M Arabic words, 3.7M Japanese words). The source texts were from a number of Arabic news agencies (e.g., Al-Ahram, Al-Hayat, Al-Nahar, Al-Quds Al-Arabi, etc.) and were translated by  undergraduate students  majoring in Arabic at TUFS.  
The authors \cite{Inoue:2018:parallel} also presented the first results on Arabic--Japanese phrase-based MT trained on their corpus. 

In this paper we make use of this corpus as part of improving the quality of Arabic-Japanese MT. 


\section{Linguistic Facts about Arabic and Japanese} 
 \label{facts}
 
 To contextualize the degree of difference between Arabic and Japanese linguistically, we compare them with each other and English in a number of dimensions (See Table~\ref{tab:ar-en-ja-table}). All Arabic examples are provided in the Buckwalter Transliteration \cite{Buckwalter:2004:buckwalter}.
 
 The most obvious differences are in the orthography: Arabic is written using the Arabic script in a highly ambiguous Abjad  orthography that omits short vowels and allows for a large number of clitics;
 English is written in the Latin Script Alphabet; and Japanese uses three different scripts together without ``word'' space -- the logographic Kanji based on Chinese, plus the Hiragana syllabary for grammatical units and basic phonology, and Katakana syllabary for foreign names and concepts.  
 Morphologically speaking, English is the poorest and Arabic the richest of these three languages.  Japanese is poorer morphologically than Arabic in some dimensions such as gender and number, but it has a more complex formality system.  Japanese also has a number of grammatical particles that parallel in some cases Arabic's rich morphological cases system (which is often unwritten since it is expressed vocalically).
  In the context of MT, it is necessary to preprocess the text to create more symmetry between source and target, especially under limited data constraints  \cite{Stymne:2012:text,Inoue:2018:parallel}.   In our post-editing system, and for evaluation, we use the same consistent word tokenizer for Japanese used by \newcite{Inoue:2018:parallel} called MeCab morphological analyzer (0.996)~\cite{kudo2005mecab} with IPAdic.

\begin{table}[t!]
\begin{center}
 \scalebox{0.75}{
\begin{tabular}{c l l | c l l }
\jap{は} & (wa) & Topic & \jap{へ} & (e) & Direction \\
\jap{が} & (ga) & Subject & \jap{の} & (no) & Possesive \\
\jap{を} & (wo) & Direct Object of a Verb & \jap{も} & (mo) & Also \\
\jap{で} & (de) & Location of an Action & \jap{と} & (to) & And \\
\jap{に} & (ni) & Location & \jap{や} & (ya) & Or \\
\end{tabular}
}
\end{center}
\caption{\label{tab:JapPart} Japanese particles and their usage}
\end{table}

Syntactically,  Japanese is a left-branching head final language, following the Subject-Object-Verb sentence structure. It uses post-positional case markers to mark arguments for grammatical and semantic roles. These case markers come in the form of particles, which act as suffixes that immediately follow the modified noun, verb, or adjective. All components and most adjuncts are marked with a case-marker, which is what leads many to describe Japanese as a free word order language. As long as each component is not separated from its case-marker, it can be moved around within a sentence (with the exception of the verb which must remain in the final position). Ten frequently used particles are described in Table~\ref{tab:JapPart}.  
All nominal modifications in Japanese precede the nominal head -- adjectives, possessive and relative clauses. Arabic, in contrast, is a right-branching verb-initial where the verb position is the exact opposite of Japanese. 
Copular constructions in Arabic do not have a verb in the present tense, and all nominal modifiers follow the nominal head.
In the context of Arabic-to-Japanese MT, we expect word order and 
syntactic case marking particle generation to be especially challenging. 
\begin{table*}[t!]
\begin{center}
 \scalebox{0.9}{
\setlength{\tabcolsep}{3pt}
\begin{tabular}{c | l | c | c |cccc|c|c}
\bf Dataset  & \bf System                         &\bf RIBES  &\bf BLEU  & \bf 1-gram & \bf 2-gram &\bf 3-gram &\bf 4-gram &\bf Brevity Penalty   &\bf Length Ratio\\\hline\hline
{\devtest} & {\GoSMT}                         & 57.86                         & 11.48 & 51.20  & 17.60  & 6.80   & 2.80   & 1.00 & 1.01  \\
{\devtest} & {\GTNMT}                        & 62.19                         & 9.51  & 51.30  & 17.90  & 7.60   & 2.50   & 0.76   & 0.79  \\\hline

\end{tabular}
}
 \end{center}
\caption{\label{tab:error-results} Baseline system results.
}
\end{table*}


\section{Errors in Arabic-Japanese MT Baselines} 
\label{errors}
In this section we present the two baseline  systems we compare against:  {\GoSMT} and {\GTNMT}. 

{\bf \GoSMT} is the system described in  \newcite{Inoue:2018:parallel}.  It is the first result of an Arabic-to-Japanese phrase-based SMT system trained on {\ARJAcorpus}. 
The SMT LM was trained on the target side of {\ARJAcorpus}.  {\bf \GoSMT} represents the basic approach to MT given a corpus in a smallish specific language-pair or domain.

{\bf \GTNMT} is Google's NMT system, the details of which are not available to us; however, we expect it  
to have access to far more training  data than {\bf \GoSMT}. Google's Arabic-Japanese NMT is actually an English pivot system utilizing Arabic-English and English-Japanese NMT systems.\footnote{We also evaluated Google's phrase-based SMT system for Arabic-to-Japanese, but it performed poorly compared to the NMT system, so we do not report on it.}

Table \ref{tab:error-results} presents the results of these two systems on the development set ({\devtest}) we use in this paper.  Details on  {\devtest}  can be found in Section~\ref{experiments}.
The BLEU scores indicate that {\GoSMT} produced more accurate Japanese, outperforming {\GTNMT} by nearly 2 BLEU points. This is consistent with {\GoSMT} being trained on in-domain data unlike {\GTNMT} which was trained on general data. 
However, interestingly, the two systems produced comparable n-gram scores, indicating that the main reason for {\GTNMT} receiving a lower overall BLEU score is its larger brevity penalty. This is not surprising given that {\GTNMT} uses a neural model which typically generates less text than SMT.  In this particular data set, the difference in length between the hypothesis and gold reference is very large ($\sim$20\%).

In a sample of 20 sentences (roughly 2,000 Japanese characters) from {\GoSMT} {\devtest}'s output, we carefully observed that around 25\% of errors involve word-order, 25\% of errors involve particles (all those appearing in Table~\ref{tab:JapPart}), 12.5\% of errors involve untranslated Arabic script and the rest involve a combination of tenses, word choice and sentence fragments. 
When examining the same 20 sentences of {\GTNMT} we found that the overall fluency of the Japanese was much higher. 
This is consistent with NMT vs SMT common wisdom. It is also reflected in the higher RIBES metric score which advantages correct order.
We also noted that while {\GoSMT} leaves unknown words in their Arabic script, {\GTNMT} translates them into English. Additionally, {\GTNMT} has a tendency to drop aspects of Japanese that are more specific to the domain. 

For example, sentence endings like \jap{と述べた} (to nobeta) meaning `is what was stated' 
are often dropped from {\GTNMT} but remain in {\GoSMT}. Similarly, {\GTNMT} tends to use the informal verb endings like \jap{になる} (ni naru) whereas {\GoSMT} uses the more formal and complete ending \jap{することになっています} (suru koto ni natte imasu). While English does not have the same exact difference, it is comparable to the formality difference in the verb "check" and its implied meaning "review and look for errors". This disparity can be attributed to the different sets of training data. {\GoSMT} likely produces formalities that more closely correspond to the reference text because it was trained on in-domain data.

 In the approach we take in this paper we will post-edit with an eye toward adapting the output of {\GTNMT} using APE. 
The APE system will be trained on the output of Arabic-to-Japanese {\GTNMT}  (as source language) to reference Japanese (as target language).

\section{Automatic Post-Editing for Arabic-Japanese MT} 
\label{postedit}
\subsection{General Approach}
We aim to develop an APE system which relies on the following assumptions: 1) The availability of a parallel corpus of the desired domain, and 2) a pre-existing general-domain MT system. We run the source side of the corpus through the general MT system and then train the post editing system on the produced output as the source side and the gold reference remains at the target side.  

\subsection{System Architecture}
We develop a post-editing system for Japanese, henceforth {\MaiSMT}. We exploit {\ARJAcorpus} as the in-domain parallel corpus for Arabic-Japanese, and use Google NMT as the general MT system in our approach. The development steps are as follows:
\begin{enumerate}
\item The raw Arabic text from the in-domain corpus ({\ARJAcorpus}) is fed to the general MT system ({\GTNMT}) to produce a Japanese translation corpus. This translated corpus constitutes the source side for training, tuning and testing of {\MaiSMT}. 
\item The translated corpus and the gold Japanese reference are tokenized using MeCab morphological analyzer (0.996)~\cite{kudo2005mecab} with IPAdic for consistency.
\item We build {\MaiSMT} as a phrase-based SMT system using Moses toolkit~\cite{koehn2007moses}  with the translated text as source  and  Japanese reference as target.
\end{enumerate}

Word alignment was done by MGIZA++~\cite{gao2008parallel} with a maximum phrase size of 8. 
The grow-diag-final-and and msdbidirectional-fe options were selected for symmetrization and reordering. KenLM Toolkit~\cite{heafield2011kenlm} was used to build the 5-gram language model (LM) we adopted in our system. We use an LM that is trained on the tokenized target side of the training data \cite{Inoue:2018:parallel}. These same settings were used by \newcite{Inoue:2018:parallel} to build their phrase-based SMT  system (i.e., {\GoSMT}). 

We evaluate {\MaiSMT} by running it on the output of {\GTNMT}. We also study the effect of doing the same with the output of {\GoSMT} for completeness.


\section{Experimental Settings and Results}  
\label{experiments}

In this section, we present the experimental settings, results and analysis of the baselines and the APE for Arabic-Japanese MT.

\subsection{Data Splits} 
We use the same splits introduced by \newcite{Inoue:2018:parallel} for the training data ({\train}), tuning data ({\tune}), development testing data ({\devtest}) and blind testing data ({\blind}) as shown in Table \ref{tab:splits}.  All Japanese files were tokenized with the same tokenization scheme provided by the MeCab morphological analyzer \cite{kudo2005mecab} with IPAdic for Japanese.

\begin{table}[tbh]
\begin{center}
\begin{tabular}{l|rrr}
& Sentences &  Tokens (ar) &  Tokens (ja)\\
 \hline
{\train} & 59,238 & 2,175,438 & 3,403,244\\
{\tune} & 621 & 23,312 & 36,595\\
{\devtest} & 2,393 & 92,760 & 147,536\\
{\blind} & 2,236 & 85,940 & 144,358\\
\end{tabular}
\caption{The basic statistics of the {\ARJAcorpus} corpus splits following  \newcite{Inoue:2018:parallel}.}
\label{tab:splits}
\end{center}
\end{table}

\begin{table*}[t!]
\begin{center}
 \scalebox{0.8}{
\setlength{\tabcolsep}{3pt}
\begin{tabular}{c | l | c | c |ccccc|cc}
\bf Dataset  & \bf System                         &\bf RIBES  &\bf BLEU  & \bf 1-gram & \bf 2-gram &\bf 3-gram &\bf 4-gram &\bf Brevity Penalty   &\bf GeoMean &\bf Ratio\\\hline\hline
{\devtest} & {\GoSMT}                         & 57.86                         & 11.48 & 51.20  & 17.60  & 6.80   & 2.80   & 1.00 & 11.44   & 1.01  \\
{\devtest} &  --\textgreater { }{\MaiSMT}  & \bf 58.04                         & \bf 13.17 & 52.80  & 19.60  & 8.10   & 3.60   & 1.00 & 13.18   & 1.06  \\\hline
{\devtest} & {\GTNMT}                        & 62.19                         & 9.51  & 51.30  & 17.90  & 7.60   & 2.50   & 0.76 & 11.49   & 0.79  \\
{\devtest} &  --\textgreater { }{\MaiSMT} & \bf  64.64                         & \bf 15.35 & 56.50  & 23.60  & 11.40  & 5.80   & 0.89 & 17.23   & 0.90  \\\hline\hline
{\blind}     & {\GoSMT}                         & 56.63                         & 9.38  & 50.50  & 15.80  & 5.70   & 2.20   & 0.94 & 10.00   & 0.94  \\
{\blind}     &   --\textgreater { }{\MaiSMT}  & \bf 57.00                         & \bf 11.44 & 52.20  & 17.70  & 6.90   & 2.80   & 0.99 & 11.56   & 0.99 \\\hline
{\blind}     & {\GTNMT}                        & 61.60                         & 8.71  & 51.50  & 17.60  & 7.40   & 3.30   & 0.71 & 12.20   & 0.75  \\
{\blind}     &  --\textgreater { }{\MaiSMT} & \bf 63.28                         &  \bf 13.12 & 56.00  & 22.10  & 10.30  & 5.10   & 0.82 & 15.97   & 0.84  \\\hline
\end{tabular}
}
\end{center}
\caption{\label{tab:main-results} {\MaiSMT} Results for {\devtest} and {\blind} on {\GoSMT} and {\GTNMT}.
}
\end{table*}
\begin{table*}[h]
\begin{center}

\begin{small}
\setlength{\tabcolsep}{3pt}
 \scalebox{0.9}{
\begin{tabular}{c | l | cc |cc | cc|cc}
\bf Dataset  & \bf System                           & \bf Match Part & \bf Match Other &\bf Misalign Part &\bf Misalign Other &\bf Ins Part &\bf Del Part & \bf BLEU & \bf Max BLEU \\\hline\hline
{\devtest} & {\GoSMT}                          & 6.32 & 13.44 & 4.47 & 7.46 & 4.18 & 5.21  & 11.48 & 20.94  \\
{\devtest} &  --\textgreater { }{\MaiSMT}   & 6.65 & 14.55 & 4.73 & 8.37 & 3.59 & 6.16  & 13.17 & 23.62  \\\hline
{\devtest} & {\GTNMT}                          & 6.36 & 11.82 & 2.93 & 3.57 & 5.88 & 3.03  & 9.51 & 14.91  \\
{\devtest} &  --\textgreater { }{\MaiSMT}  & 6.91 & 15.37 & 3.48 & 5.17 & 4.75 & 4.27  & 15.35 & 22.49  \\\hline\hline
\end{tabular}
}
\caption{\label{tab:match-part-oth} Error Analysis (per sentence) on particles v. other. BLEU score taken from Table \ref{tab:main-results} for comparison. Max BLEU is the maximum BLEU possible by only allowing movements and particle insertion/deletions.
}
\end{small}
    
\end{center}
\end{table*}

\subsection{Metrics}
We used two different automated MT evaluation metrics to quantitatively evaluate our outputs: BLEU \cite{Papineni:2002:bleu} and RIBES \cite{Isozaki:2010:automatic}. 
We used the  Moses toolkit  implementation for BLEU scores. For  RIBES, we used  version (1.03.1.) with the default $\alpha$=0.25 and $\beta$=1.0 values.\footnote{\url{http://www.kecl.ntt.co.jp/icl/lirg/ribes/}} The two metrics are calculated based on different factors and together provide us with a more comprehensive metric. BLEU is a precision-base metric that is found by summing the n-gram matches for every predicted sentence in the corpus. To emulate recall, a brevity penalty (BP) is introduced to compensate for the possibility of high precision translations that are much shorter than the reference text.

Though BLEU provides a very good estimate for the similarity of a text to its reference, its n-gram-based method lacks explicit consideration of re-ordering beyond a small window, giving very little penalty to texts with low tri/four-grams. RIBES compensates for this disadvantage by adding a rank correlation coefficient prior to unigram matches without the need for higher order n-gram matches. This produces a metric that takes into account re-ordering, which serves to be quite useful information in our research. Together, BLEU and RIBES provide an accurate measurement for determining the success of the translations.

\subsection{Experimental Results}  
We conducted a number of experiments, which evaluate the performance of {\GoSMT} and {\GTNMT} on {\devtest} and {\blind}, then the performance of our {\MaiSMT} on top of them, in terms of RIBES and BLEU scores. Table \ref{tab:main-results} presents these scores in addition to a number of internal BLEU scores: the 1~to~4 n-gram precision, brevity penalty, geometric mean (GeoMean) of the n-grams (basically BLEU without the brevity penalty) and the hypothesis to reference length ratio.

\begin{table*}[t]

 \begin{center}
 \scalebox{0.9}{
 \begin{small}
\begin{tabular}{p{0.09in} p{0.9in} p{5.5in}}
\bf (a) &\bf Arabic Input & \multicolumn{1}{r}{<w {\lq}kryAt 'rb`{\rq} m`rwfT bAlt.trf.>} \\ 
    & & \multicolumn{1}{r}{\it And Kiryat Arba is known for its extremism.} \\\hline
 & \bf Gold Reference  & \jap{「キルヤト・アルバア」は過激だとして知られている。} \\
& & {\it Kiryat Arba [topic] radical is known.} \\\hline
& {\GTNMT} & \jap{そして「KiryatArba」は過激主義として知られています。} \\
 & & {\it And Kiryat Arba [topic] radical extremist is known.} \\\hline
& ++{\MaiSMT} & \jap{そして「「カリヤート・アルバア」は、過激主義として知られている。}\\
& & {\it And Kiryat Arba [topic] radical extremist is known.}\\\hline\hline

\bf (b) &\bf Arabic Input & \multicolumn{1}{r}{<kAn Alra'iys wAlwfd Alm.srY qd w.slwA 'ilY Al`A.smT Als`wdyT .zhr 'ms.> }\\
    & & \multicolumn{1}{r}{\it The Egyptian president and delegation arrived to the Saudi capital at noon yesterday.} \\\hline

 & \bf Gold Reference & \jap{スィースィー大統領とエジプト代表団は昨日の昼にリヤドに到着した。} \\
 & & {\it Sisi President and Egypt delegation [topic] yesterday [poss] afternoon in Riyad [location] arrived.}\\\hline
& {\GTNMT} & \jap{大統領とエジプト代表団は昨日の午後サウジの首都に到着した。}\\
& & {\it President and Egypt delegation [topic] yesterday [poss] afternoon Saudi [poss] capital [location] arrived.}\\\hline
& ++{\MaiSMT} &\jap{大統領とエジプト側の代表団は昨日(21日)の午後にサウジアラビアの首都に到着した。}\\
& & {\it President and Egypt-side [poss] delegation [topic] yesterday (the 21st) [poss] afternoon in Saudi Arabia [poss] capital [location] arrived.}\\\hline\hline

\bf (c) &\bf Arabic Input & \multicolumn{1}{r}{<wbl.gt mwAznT AldwlT lil`Am Al.hAly 20,7 blywn dynAr (68,1 blywn dwlAr).>}\\
    & & \multicolumn{1}{r}{\it The national budget for this year has reached 20.7 billion dinar (68.1 billion dollars).} \\\hline

 & \bf Gold Reference & \jap{今年度の国家予算は、207億ディナール(681億ドル)に達していた。} \\
 & & {\it This year [poss] national budget, 207 oku ($10^8$) dinars (681 oku ($10^8$) dollars) [location] reached.}\\\hline
& {\GTNMT} & \jap{今年度の州予算は207億ドル（681億 ドル）に達しました。}\\
& & {\it This year [poss] state budget [topic] 207 oku ($10^8$) dollars (681 oku ($10^8$) dollars) [location] reached.} \\\hline
& ++{\MaiSMT} &\jap{今年度の国家予算は207ディルハム(681 ドル)に達した。}\\
& & {\it This year [poss] national budget [topic] 207 Dirhams (681 dollars) [location] reached.}\\\hline\hline

\bf (d) &\bf Arabic Input & \multicolumn{1}{r}{<w_dlk li-'i`.tA' Al-'urdn AlqdrT m^gdadA `lY mumArsT dwrh AltAry_hy fy Alq.dyT Alfls.tynyT.>}\\
    & & \multicolumn{1}{r}{\it In order to renew Jordan's ability to perform its historic role in the Palestine problem.} \\\hline

 & \bf Gold Reference & \jap{協定の見直しは、パレスチナ 問題においてヨルダンに歴史的役割を果たす力を与える。} \\
 & & {\it Agreement [poss] review [topic], Palestine problem [in] Jordan [in] historical role [directobj] power [directobj] gives.} \\\hline
& {\GTNMT} & \jap{ヨルダンにパレスチナの原因においてその歴史的な役割を果たす能力を与えるために。}\\
& & {\it Jordan [location] Palestine [poss] cause that historical role [directobj] ability [directobj] gives.}\\\hline
& ++{\MaiSMT} &\jap{ヨルダンにパレスチナ問題に関しては、その歴史的役割を果たす能力を与えるために。}\\

 & & {\it Jordan [location] Palestine problem about [topic], that historical role [directobj] ability [directobj] gives.} \\\hline
 
\end{tabular}
\end{small}
}
\end{center}

\caption{\label{tab:output-examples} Examples from {\devtest} of Arabic, Gold Japanese,  {\GTNMT}, and {\GTNMT}++{\MaiSMT}  systems. The English text below the Japanese is a word-by-word gloss.
}
\end{table*}


{\MaiSMT} improves upon the baseline systems for both {\devtest} and {\blind} on both BLEU and RIBES.
We used the bootstrap resampling method\footnote{bootstrap-hypothesis-difference-significance.pl script provided as part of Moses toolkit.} implemented by \newcite{Koehn:2004:statistical} to compute the statistical significance of the empirical results comparing the baseline systems to {\MaiSMT}, assuring that the improvements are real. We achieved $p$-value $<0.05$ in all comparisons, indicating the significance of all the improvements we attained.  

The absolute increase in BLEU and RIBES for the {\GTNMT} is bigger than the increase for {\GoSMT}. This is understandable given that {\MaiSMT} was designed for post-editing {\GTNMT}. It is still interesting that we see some statistically significant increase in {\GoSMT} post-editing suggesting that the {\MaiSMT} is addressing some shared error phenomena in both systems. 
%
Focusing on {\GTNMT}'s postedit, we note a BLEU score increase of 5.84\% absolute on the {\devtest}.  This increase comes from three sources: First, a basic increase in unigram precision (5.20\%) which is likely to be the result of correctly added words as well as corrected word. Second, an average of (4.27\%) increase in the 2,~3,~and~4-gram precision scores, which is connected to better reordering. Third, the brevity penalty effect is reduced by 54\% (from a multiplier of 0.76 to 0.89), which is the result of added words alone.
The results on the {\blind} are comparable but slightly lower.
In the next section, we present an error analysis and examples to help us understand these changes further.


\section{Error Analysis and Examples}  
\label{analysis}

\subsection{Error Analysis}
Inspired by tools for automatic MT error analysis such as Blast \cite{Stymne:2011:blast} and Ameana \cite{ElKholy:2011:automatic}, we developed a Japanese-targeted automatic error analysis system.  The insights and motivation for this system came from our initial study of errors in Japanese output of MT (Section~\ref{errors}).
The system targets two primary errors in the Japanese output: 1) incorrect placement of correctly produced tokens and 2) incorrect particle usage. Given a predicted output and its gold reference, we use a dynamic programming algorithm to create the optimal alignments of the two sentences, requiring the minimum number of insert/delete (but no movement or substitution) edits to transform the translation into the reference. We then allow three transformations on the alignments: 1)~movements, 2)~particle insertion and 3)~particle deletion, to produce the best version of the predicted translation through doing these operations only.  Movements are allowed for all words (i.e., including particles). The only particles we allow inserting or deleting are those in Table~\ref{tab:JapPart}. 
Based on the alignment and transformation, we produce a number of scores for the whole document, keeping a separation into ``particles'' and ``other'' (non-particle words).
The scores identify the average per sentence for:
\begin{itemize}
    \item Direct particle matches
    \item Direct other matches
    \item Misaligned (i.e. moved) particles matches
    \item Misaligned other matches
    \item Inserted particles
    \item Deleted particles
\end{itemize}

All of these scores are shown for the {\devtest} in Table~\ref{tab:match-part-oth}.  The table also adds two scores: the BLEU score (which matches that in Table~\ref{tab:main-results}); 
and an oracular Max BLEU computed  with the transformed Japanese sentences against the gold references.
Although oracular and very generous, the Max BLEU score gives us a sense of the maximum possible score that can be attained using the limited operations of movement and particle insertion/deletion.


Across both baseline systems,  {\MaiSMT} produces a higher number of correctly matched words; this is consistent with the unigram precision increase shown in Table~\ref{tab:main-results}.  The increase of matches, aligned and misaligned is in comparable relative proportions in the baselines and the post-edited version: For {\GTNMT}, the misaligned matches are $\approx$27\% of all matches for the baseline and post-edited versions. The respective ratio for {\GoSMT} is higher at $\approx$38\%.

The needed particle insertions in the post-edited version are lower than the baseline; but the particle deletions are higher. The fewer insertions suggest that post-editing is adding correct particle; however, the higher deletions may be linked to either superfluous particles or incorrect particle insertions -- i.e., a particle was inserted but had to be substituted for another particle.

The Max BLEU score for the {\GTNMT} baseline is lower than the BLEU score for the post-edited version. This is explainable by the fact that the Max BLEU computation does not account for insertion of non-particle words or word substitutions which happen during post-editing.  It is interesting to note that the Max BLEU for the {\GoSMT} baseline was far from  reached after post-editing. 
The Max BLEU score for {\GTNMT}$\rightarrow${\MaiSMT} suggest a lot of potential improvements may be possible just with movement and particle insertion, deletion, and substitution.

\subsection{Example Translations}
Table \ref{tab:output-examples} shows four representative sentence-level examples of our improvements to {\GTNMT}'s Japanese. We discuss a number of specific phenomena across these examples below.

\paragraph{Katakana Changes} Sentence (a) shows the ability of {\MaiSMT} to convert English to Katakana. While Katakana is widely used for proper nouns adopted from other languages, it requires the ability to transliterate the foreign word. This can be difficult in Arabic that omits short vowels and thus leaves pronunciation ambiguous to a non-speaker. English-Katakana is slightly easier given the abundance of English vowels, but {\GTNMT}'s output highlights how it is still difficult. {\MaiSMT} shows the ability to convert this English into an acceptable transliteration of Japanese. While the Katakana is not identical, it better captures the meaning than English text. Given the limitation of the approach we use, all of the successful changes made had to be present in the training data.

\paragraph{Particle Changes} Sentence (b) shows the ability of {\MaiSMT} to correctly insert missing particles. {\GTNMT} output includes [yesterday afternoon] but fails to add the particle \jap{に} (ni `in') following \jap{昨日の午後} (kinou no gogo `yesterday afternoon') which signals that the action occurred in the specific time-frame. Sentence (d) shows another instance of correcting suffixes when {\MaiSMT} is able to remove the unnecessary \jap{な} (na `adjectival ending') following  \jap{歴史的} (rekishiteki `historic') which is absent in the gold reference.  The deletion of  \jap{な} (na) has no effect on  neither the adequacy nor fluency, but it reflects a closer adaptation to the style of the corpus we use.

\paragraph{Word Changes}  Sentence (c) shows two instances of incorrect word choice that {\MaiSMT} is able to handle correctly. {\GTNMT} incorrectly translates Arabic <mwAznT AldwlT> {\it mwAznp Aldwlp} `national budget' to `state budget' of which {\MaiSMT} correctly translates to `national budget'. The other incorrect choice of words by {\GTNMT} is the first instance of the currency `dollar' which should be `dinars' according to the gold reference. {\MaiSMT} is able to capture this error and then replace 
it with `dirhams' instead. While this does not match the gold reference, it produces a sentence that understands the disparity in currency. Another instance of correcting word choice can be found in sentence (d) when {\MaiSMT} changes the word `cause' (\jap{原因} (genin)) to `problem'(\jap{問題} (mondai)). While both are grammatically correct, the word `problem' (\jap{問題} (mondai)) is more often used in discussing matters such as the Palestinian conflict. This word change is considered an appropriate change that corresponds with the gold reference and the domain of the corpus we use.

Though the overall improvement of BLEU and RIBES scores indicates a closer match to the Gold reference text, these sentence level corrections prove that {\MaiSMT}'s improvements correlate with the fluency of the Japanese output. 

\section{Conclusion and Future Work}
In this paper, we presented results from an automatic post-editing system for Arabic-Japanese MT that exploits a corpus of Arabic-Japanese news articles. Our results improve over a very strong (but domain/genre agnostic) state-of-the-art system.
%
A detailed linguistic analysis provided some insights in the type of changes inflected by the post-editing system. 
 
As future work, we want to explore the possibility of incorporating source-language-specific information into the post-editing system to allow for more consistent translations with the initial Arabic source text. The oracular Max BLEU scores we calculated suggest that there is still some room for improvement just within the the space of reordering and particle insertion/deletion.


\bibliographystyle{mtsummit2019}

\end{document}